%% file: main_camera_ready.tex
\documentclass[10pt,twocolumn,letterpaper]{article}

\usepackage{cvpr}              % To produce the CAMERA-READY version
\usepackage{float}
\usepackage{cuted}
\usepackage{graphicx}
\usepackage{colortbl}
\usepackage{multirow}
\usepackage{booktabs}
\usepackage{siunitx}

\definecolor{cvprblue}{rgb}{0.21,0.49,0.74}
\usepackage[pagebackref,breaklinks,colorlinks,allcolors=cvprblue]{hyperref}
\usepackage[table]{xcolor}
\usepackage{diagbox}
\usepackage{ulem}
\def\paperID{30762} % *** Enter the Paper ID here
\def\confName{CVPR}
\def\confYear{2026}
\usepackage{xcolor}  % 引入颜色宏包
\usepackage[accsupp]{axessibility}

\title{
3M-TI: High-Quality Mobile Thermal Imaging \\ via Calibration-free Multi-Camera Cross-Modal Diffusion 
}

\author{
Minchong Chen$^{1,2,\dagger}$ \quad 
Xiaoyun Yuan$^{1,2,\dagger,*}$\quad 
Junzhe Wan$^{1}$ \quad 
Jianing Zhang$^{3,4}$ \quad 
Jun Zhang$^{4}$ \\
$^{1}$MoE Key Lab of Artificial Intelligence, AI Institute, Shanghai Jiao Tong University \\
$^{2}$School of Computer Science, Shanghai Jiao Tong University \quad
$^{3}$Fudan University \quad
$^{4}$Tsinghua University \\
{\tt\small yuanxiaoyun@sjtu.edu.cn}
}

\begin{document}
\maketitle

\begin{strip}
\begin{minipage}{\textwidth}
\vspace{-30pt}
\centering
\includegraphics[width=\textwidth]{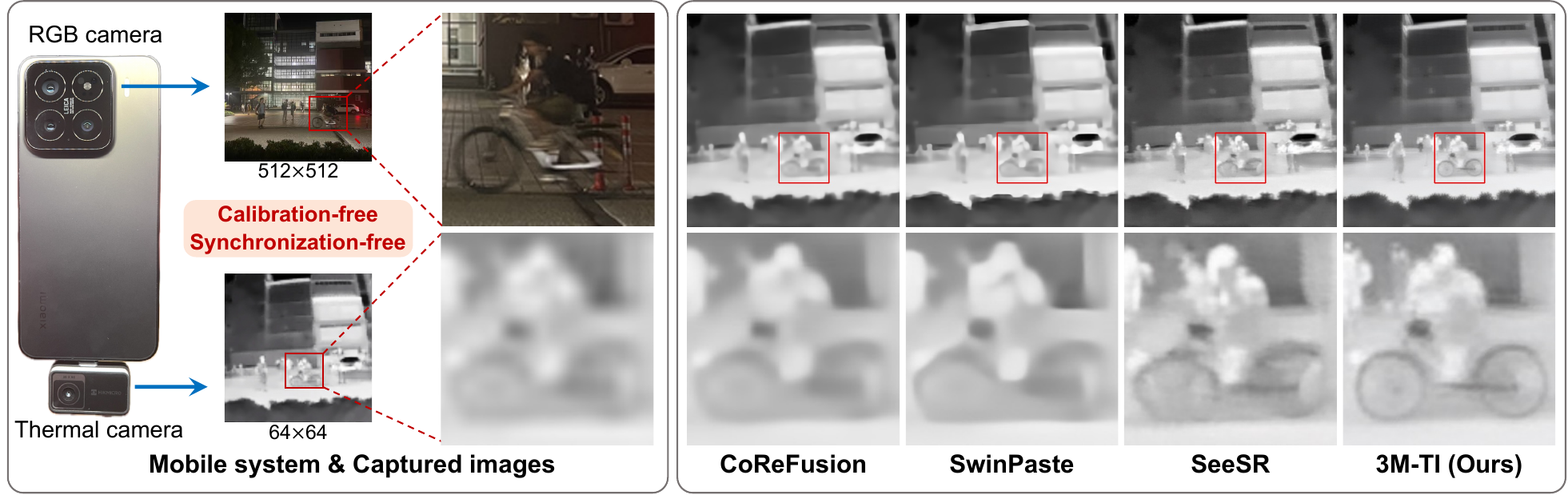}
\captionof{figure}{
A smartphone-based mobile imaging system integrating calibration-free and synchronization-free RGB and thermal cameras. The proposed 3M-TI method delivers superior thermal image quality compared with state-of-the-art restoration approaches.}
\label{fig:teaser}
\vspace{-5pt}
\end{minipage}
\end{strip}
\renewcommand{\thefootnote}{}
\footnotetext[0]{$^\dagger$ These authors have contributed equally to this work.}
\footnotetext[0]{* Corresponding author. Mail: yuanxiaoyun@sjtu.edu.cn.}

% \begin{figure*}[t]
%     \centering 
%     \includegraphics[width=\textwidth]{sec/abstract.png}
%     \caption{Comparison results on images captured by a little thermal camera. Compared to state-of-the-art restoration methods, our approach achieves superior reconstruction quality and visual perception.}
%     \label{fig:abstract}
% \end{figure*}

\input{sec/0_abstract}    
\input{sec/1_intro}
\input{sec/2_related}
\input{sec/3_method}

\input{sec/4_experiment}
\input{sec/5_conclusion}
{
    \small
    \bibliographystyle{ieeenat_fullname}
    \bibliography{reference}
}

\end{document}

%% file: sec/0_abstract.tex
\begin{abstract}
The miniaturization of thermal sensors for mobile platforms inherently limits their spatial resolution and textural fidelity, leading to blurry and less informative images.
Existing thermal super-resolution (SR) methods can be grouped into single-image and RGB-guided approaches: the former struggles to recover fine structures from limited information, while the latter relies on accurate and laborious cross-camera calibration, which hinders practical deployment and robustness.
Here, we propose 3M-TI, a calibration-free Multi-camera cross-Modality diffusion framework for Mobile Thermal Imaging. 
At its core, 3M-TI integrates a cross-modal self-attention module (CSM) into the diffusion UNet, replacing the original self-attention layers to adaptively align thermal and RGB features throughout the denoising process, without requiring explicit camera calibration.
This design enables the diffusion network to leverage its generative prior to enhance spatial resolution, structural fidelity, and texture detail in the super-resolved thermal images.
Extensive evaluations on real-world mobile thermal cameras and public benchmarks validate our superior performance, achieving state-of-the-art results in both visual quality and quantitative metrics. More importantly, the thermal images enhanced by 3M-TI lead to substantial gains in critical downstream tasks like object detection and segmentation, underscoring its practical value for robust mobile thermal perception systems. Project page: \url{https://github.com/work-submit/3MTI}
\end{abstract}

%% file: sec/1_intro.tex
\vspace{-15pt}
\section{Introduction}
\label{sec:intro}
Thermal imaging detects infrared radiation emitted by objects, extending perception beyond the visible spectrum for enhanced scene understanding.
It is particularly effective under challenging conditions such as darkness, fog, or smoke, providing reliable scene information that complements RGB imaging for multimodal perception and fusion \cite{liu2025dcevo,ma2019infrared,tang2023divfusion}.
Owing to its robustness, thermal imaging has become increasingly valuable in safety-critical applications, such as autonomous driving, robotic navigation, and situational awareness \cite{shin2023deep,tang2023happened,sheinin2024projecting}.

Despite rapid advances in infrared sensing, mobile thermal imaging remains fundamentally constrained by hardware. Miniaturization for compact devices reduces aperture size, lowering spatial resolution and textural detail. In addition, the long wavelength of thermal radiation limits the minimum pixel size, while the high cost of infrared sensors further restricts pixel counts compared with RGB cameras.
These factors jointly result in blurry and less informative thermal images on mobile platforms, motivating extensive efforts to enhance their resolution and perceptual quality through computational reconstruction.

A representative example of such computational reconstruction is thermal image super-resolution (SR), which aims to reconstruct high-resolution spatial details from low-resolution thermal observations \cite{wang2020deep}. 
However, the inherently limited information in low-resolution observations, aggravated by the lack of large-scale thermal datasets, makes thermal super-resolution highly underconstrained and hinders network training and generalization compared with RGB-based tasks.
To compensate for this deficiency, RGB-guided SR methods exploit cross-modal cues to enhance reconstruction \cite{cnncsr,corefusion,swinfusr}.
Nevertheless, due to the fundamental differences in imaging principles, thermal and RGB images often exhibit substantial  discrepancies, and directly merging their features may introduce unrealistic details \cite{dcevo,Metafusion,zheng2024probing}.
Moreover, most existing methods rely on pixel-level aligned RGB-thermal pairs, requiring laborious calibration, which severely limits robustness and scalability in practical deployment.

% Motivation and methodology of our method.
To address these challenges, we propose 3M-TI, a calibration-free \textbf{M}ulti-camera cross-\textbf{M}odal diffusion framework for \textbf{M}obile \textbf{T}hermal \textbf{I}maging. % high-resolution thermal imaging.
Firstly, rather than relying on pixel-level calibration between RGB and thermal cameras, 3M-TI performs alignment in the latent space of a variational autoencoder (VAE) \cite{VAE}, where continuous and disentangled representations facilitate robust cross-modal correspondence. An attention-based fusion module further integrates RGB and thermal representations in this latent domain, achieving reliable alignment and fusion without explicit registration.
Secondly, because existing RGB–thermal datasets are typically small, homogeneous, and strictly aligned \cite{LLVIP,M3FD}, we design a data augmentation strategy that intentionally offsets RGB–thermal pairs to simulate real-world conditions.
This encourages the attention module to learn robust cross-modal correspondence and fusion without relying on explicit pixel-level alignment.
Furthermore, we incorporate a one-step latent-space diffusion \cite{ADD} to suppress fusion artifacts, enhance visual fidelity, and improve robustness under limited-data conditions.

Finally, we validate the proposed 3M-TI framework on a real mobile thermal imaging system. Experiments show that it significantly outperforms existing state-of-the-art methods, highlighting both its effectiveness and practical potential.
The main contributions of 3M-TI can be summarized as follows:
\begin{itemize}
\item \textbf{Calibration-free Fusion}: Fully automatic alignment and fusion of uncalibrated RGB–thermal pairs via a cross-modal self-attention module, without requiring pixel-level image registration.
\item \textbf{Misalignment Augmentation}: Camera-pose augmentation that generates intentionally misaligned RGB counterparts, enabling the cross-modal self-attention module to handle multi-camera parallax and unsynchronized captures.
\item \textbf{Cross-modal Diffusion}: Integration of latent-space diffusion to suppress fusion artifacts, enhance detail realism, and leverage pretrained priors to compensate for limited thermal data.
\item \textbf{Practical Validation}: Evaluation on both public datasets and a real mobile thermal imaging system, demonstrating effectiveness and practical potential.
\end{itemize}

%% file: sec/2_related.tex
\vspace{-6pt}
\section{Related Work}
\label{sec:related_work}
Computational thermal imaging intersects several areas of computer vision and image processing. In this section, we review the most relevant work in three directions: classical and learning-based image restoration, reference-guided image super-resolution, and recent diffusion-based approaches.

\subsection{Image Restoration}
Image restoration seeks to recover a high-quality image from a degraded observation by inverting an imaging model. 
Classical image restoration methods are typically model-based, formulated as inverse problems with explicit priors.
Variational methods regularize the solution with smoothness or edge-preserving constraints \cite{rudin1992nonlinear}.
Sparse representation approaches exploit dictionary learning and sparsity priors to recover fine details \cite{mairal2007sparse,dong2012nonlocally}.
Non-local self-similarity models, such as non-local means \cite{buades2011non} and BM3D \cite{dabov2007image}, leverage the redundancy of similar patches across the image for collaborative filtering.
Deep learning has fundamentally reshaped image restoration by learning powerful data-driven priors directly from large-scale datasets. CNN-based models (e.g., SRCNN\cite{srcnn}, EDSR\cite{lim2017enhanced}) pioneered end-to-end restoration pipelines, while recent Transformer architectures such as SwinIR\cite{swinir} further improved long-range dependency modeling. These advances have led to state-of-the-art performance in denoising \cite{swinir,li2023efficient,ipt}, deblurring \cite{kong2023efficient,liu2024deblurdinat}, and super-resolution \cite{srcnn,swinir,chen2023activating,chen2023dual,srdiff}.
Nevertheless, single-image restoration remains challenging under severe degradation or large upscaling factors, due to the limited high-frequency information in a single observation, motivating approaches that leverage additional cues.

\subsection{Reference-Guided Image Restoration}
Reference-guided image restoration leverages an auxiliary high-resolution modality (typically RGB) to recover fine details lost in degraded observations. This paradigm has been explored across multiple tasks, including RGB image super-resolution \cite{zheng2018crossnet,tan2020crossnet++,zhang2019image,fang2023engram}, hyperspectral image super-resolution \cite{cnncsr,SSC-HSR}, and thermal image enhancement \cite{swinfusr,swinpaste}.
Early approaches, such as CoReFusion \cite{corefusion}, employ a dual-encoder UNet to jointly extract and fuse RGB–thermal features. A contrastive loss is further introduced to improve cross-modal consistency and overall restoration quality.
With the advent of Transformers \cite{attention}, attention-based fusion has become the dominant paradigm. MGNet \cite{MGNet} mines multi-level cues (appearance, edge, and semantics) from RGB guidance to enhance UAV thermal super-resolution. SwinFuSR \cite{swinfusr} introduces a lightweight Swin-Transformer backbone with robust training under missing guidance, and SwinPaste \cite{swinpaste} further improves it via data mixing and multi-scale supervision. MSFFCT \cite{MSFFCT} combines channel-based Transformers with multi-scale feature fusion to capture long-range dependencies and rich thermal-RGB correlations.
However, the large domain shift between RGB and thermal images poses a significant challenge to establishing reliable cross-modal alignment, especially under uncalibrated conditions.

\subsection{Diffusion-Powered Image Restoration}
Diffusion models have recently demonstrated remarkable capabilities in image restoration and super-resolution by learning powerful generative priors that capture rich textures and structural details \cite{diffusion}.
This generative nature allows diffusion models to recover fine structures and high-frequency details that are challenging for CNN- and Transformer-based approaches.
However, the original diffusion model is computationally expensive due to its long iterative denoising process, limiting practical deployment.
To address this, several efficient variants have been proposed: latent-space diffusion methods \cite{dmdiff,seesr,sinsr,osediff} perform denoising within the compact latent representation of a VAE, reducing computation overhead while preserving high-fidelity reconstructions; distillation-based approaches \cite{huang2023knowledge,meng2023distillation,ADD} transfer the generative ability of a large teacher model to a lightweight student, enabling faster inference with minimal quality loss.

Diffusion models have also been applied to thermal image super-resolution \cite{cortes2024exploring,difiisr}.
DifIISR \cite{difiisr} incorporates gradient-based priors and a thermal spectrum distribution regulation into the diffusion process, guiding the model to capture the unique frequency characteristics of infrared images and achieving strong visual quality and downstream task performance.
\cite{cortes2024exploring} adapts the ResShift diffusion model\cite{resshift} to thermal imaging with an uncertainty-aware approach. This method produces a confidence map that evaluates the reliability of each reconstructed region, thereby enhancing pixel-level fidelity and facilitating texture reconstruction.

However, existing methods still rely on strictly pixel-aligned datasets and explicit calibration between cameras, making them sensitive to spatial and temporal misalignment and difficult to generalize to real-world mobile scenarios.
These limitations motivate our design of 3M-TI, a calibration-free latent-space diffusion framework that leverages multi-modal information for robust mobile thermal imaging.

%% file: sec/3_method.tex
\vspace{-6pt}
\section{Method}
\label{sec:method}
\begin{figure*}[htbp]
    \centering 
    \includegraphics[width=0.85\textwidth]{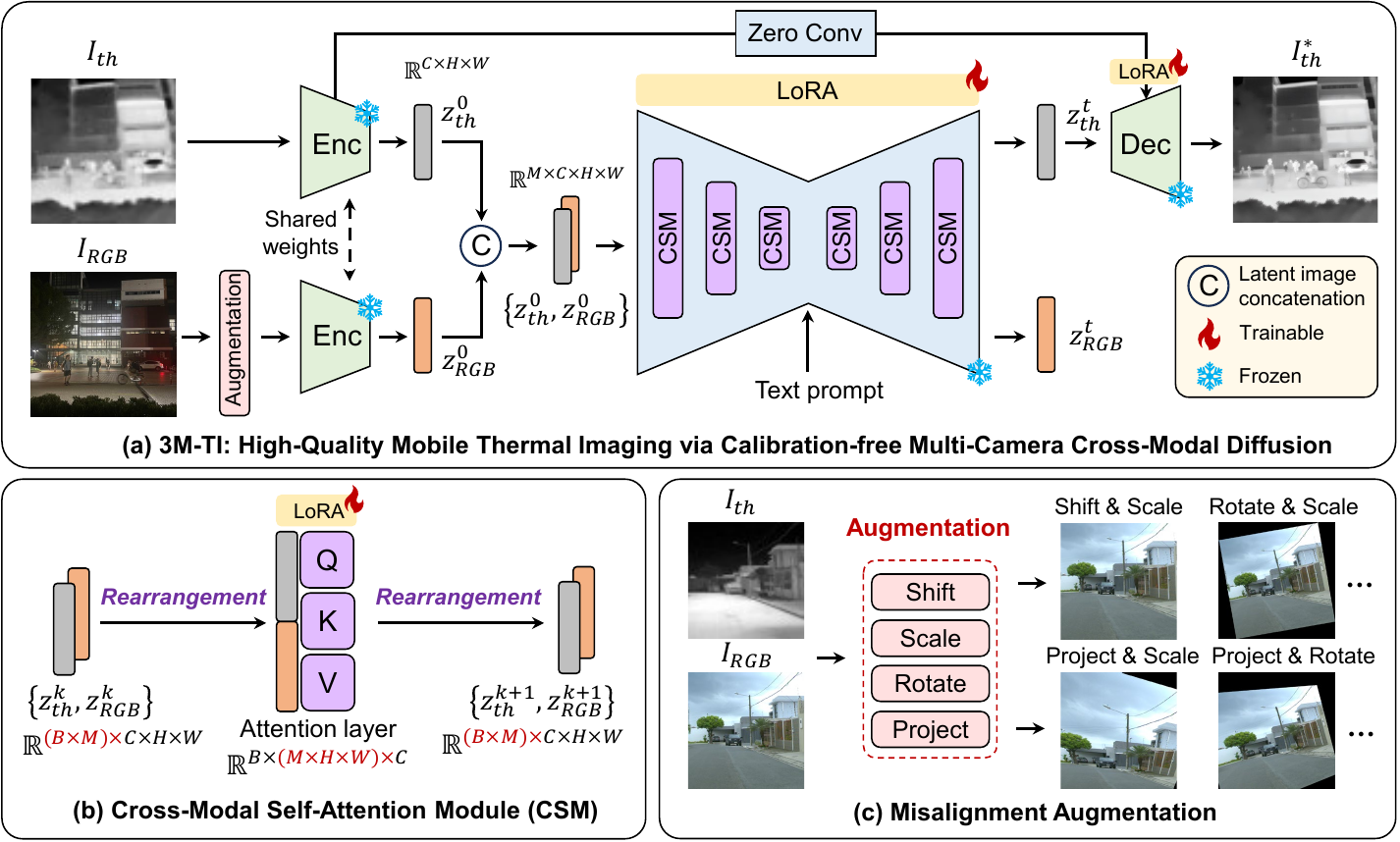}
    \caption{Overview of the 3M-TI architecture. (a) 3M-TI framework. The core of 3M-TI is a one-step diffusion-based model equipped with a cross-modal self-attention module (CSM) and a misalignment augmentation strategy. LoRA fine-tuning is applied to both the UNet and the VAE decoder. (b) Cross-modal self-attention module (CSM). Two rearrangement layers are inserted before and after the original self-attention layers to capture cross-modal correspondences. (c) Misalignment augmentation. A data augmentation strategy designed to enhance model robustness against camera parallax and temporal misalignment between RGB and thermal inputs.}
    \label{fig:method_framework}
\end{figure*}

We propose 3M-TI, a cross-modal diffusion frame, for high-quality thermal image reconstruction from uncalibrated RGB-thermal image pairs (Fig.~\ref{fig:method_framework}). 3M-TI specifically addresses three key challenges: (1) uncalibrated and unsynchronized RGB-thermal captures, (2) fusion of heterogeneous RGB-thermal representations, and (3) limited data scale and diversity of RGB-thermal datasets.

Section~\ref{overview} provides an overview of the 3M-TI framework.
Section~\ref{sec:csa} introduces the cross-modal self-attention (CSA) module, which performs automatic alignment and fusion within the VAE latent space.
Section~\ref{sec:augmentation} presents a misalignment augmentation strategy that simulates realistic inter-camera parallax and temporal offset to improve the robustness of CSA.
Finally, Section~\ref{sec:imaging_system} details the practical implementation and validation of 3M-TI on a smartphone-based multi-camera thermal imaging system.

\subsection{3M-TI Framework Overview}
\label{overview}
The 3M-TI framework reconstructs a high-resolution thermal image from a low-resolution thermal input and an uncalibrated high-resolution RGB reference, as illustrated in Fig.~\ref{fig:method_framework}(a). Our approach is built upon the one-step diffusion model, SD-Turbo~\cite{ADD}, to ensure efficient inference. The process begins by encoding the thermal and RGB images into latent representations using the frozen VAE encoder.
To establish cross-modal correspondence and fusion in the latent space, we introduce the cross-modal self-attention module (CSM). This module replaces the original self-attention layers in the diffusion UNet with our cross-modal self-attention layers, which are designed to learn multiscale correspondences between the RGB and thermal latents (Fig.~\ref{fig:method_framework}(b)). During training, we apply misalignment augmentation to the RGB images to enhance robustness against uncalibration and temporal unsynchronization (Fig.~\ref{fig:method_framework}(c)). 
Furthermore, a skip connection~\cite{parmar2024one} is incorporated to enhance structural consistency and mitigate geometric distortions.
Specifically, the feature maps from 4 encoder downsampling blocks are passed through a \(1\times1\) zero-initialized convolutional layer before being added to the corresponding decoder upsampling blocks.
Since extracting reliable semantics directly from low-resolution thermal images is suboptimal, we generate text prompts from the corresponding RGB images using the Recognize Anything Model (RAM)~\cite{ram}. Finally, we employ low-rank adaptation (LoRA)~\cite{lora} to fine-tune both the UNet and the VAE decoder.

\subsection{Cross-Modal Self-Attention Module}
\label{sec:csa}
To establish cross-modal correspondence between uncalibrated RGB and thermal images, we introduce the cross-modal self-attention module (CSM). Inspired by video- and multi-view diffusion models \cite{svd, difix}, our key idea is to repurpose the transformer blocks within the pretrained diffusion UNet for cross-modal RGB-thermal correspondence and fusion.
The RGB and thermal inputs are first encoded into a concatenated latent tensor \(\{z^0_{RGB}, z^0_{th}\} \in \mathbb{R}^{B \times M \times C \times H \times W}\), where \(B\) is the batch size, \(M=2\) denotes the number of images (RGB and thermal, can be extended to multiple RGB images), and \(C, H, W\) are the channel and spatial dimensions. 

For non-transformer blocks in the diffusion UNet, the image-number dimension \(M\) is folded into the batch dimension so that RGB and thermal images are processed independently.
Within the transformer blocks (Fig.~\ref{fig:method_framework}b), two tensor rearrangements are applied. Before entering a block, the latent tensor is reshaped to \(\mathbb{R}^{B \times (M \times H \times W) \times C}\), treating each pixel as a token and merging all pixels from both modalities into a single sequence with a length of \(M \times H \times W\). Self-attention then computes pairwise dependencies among all RGB and thermal pixels, naturally enabling the transformer to learn cross-modal correspondences and integrate complementary cues.
After the block, the latent tensor is reshaped back to \(\mathbb{R}^{(B \times M) \times C \times H \times W}\) for subsequent layers. 

At the output stage, the refined thermal latent is merged with the initial latent in the VAE encoder via a zero-initialized skip connection, producing the final high-fidelity thermal image.
The cross-modal self-attention module introduces no additional parameters, allowing the UNet and VAE decoder to be efficiently fine-tuned using LoRA \cite{lora}.

\subsection{Misalignment Augmentation}
\label{sec:augmentation} 
Existing RGB-thermal datasets are typically small and strictly pixel-aligned, limiting generalization to practical multi-camera configurations. 
In practice, spatial misalignment arise from camera parallax and temporal unsynchronization: parallax varies with scene depth and baseline, while temporal offsets are determined by object motion and capture delay.

Since our goal is to learn a model that is robust to these geometric variations, rather than overfitting to a specific calibrated setup, we propose a misalignment augmentation strategy without physical simulation. 
3M-TI applies a set of controlled spatial transformations to the RGB images, including translation, scaling, rotation, and perspective warping. The transformation parameters are chosen to reflect typical deviations encountered in handheld and multi-camera scenarios (see Figure~\ref{fig:method_framework}(c)).

This misalignment augmentation encourages the CSM to learn robust cross-modal correspondences under uncalibrated and unsynchronized conditions, effectively bridging the gap between constrained training data and diverse deployment environments.

\subsection{Practical Multi-Camera System}
\label{sec:imaging_system}
To validate our method under realistic conditions, we constructed a multi-camera system comprising a HIKVISION P09 thermal camera module (costing less than 100 US Dollars) and a Xiaomi 15 smartphone, connected via a Type-C interface (Fig.~\ref{fig:teaser}).
The thermal module provides a native resolution of \(96 \times 96\), which is resized to \(64 \times 64\) in this work, with a \(\ang{50} \times \ang{50}\) field of view (FOV) and a pixel pitch of 12~µm.
For the RGB reference, we utilize the primary camera of the Xiaomi 15, which is equipped with an OV50H sensor and captures images at a resolution of \(4096 \times 3072\). This RGB camera has an equivalent focal length of 23 mm, yielding a FOV of \(\ang{74} \times \ang{59}\), slightly wider than that of the thermal module.
This system offers a realistic platform to evaluate the performance of 3M-TI under practical, uncalibrated mobile imaging conditions.

%% file: sec/4_experiment.tex
% \clearpage
% \setcounter{page}{1}
% \maketitlesupplementary
\vspace{-6pt}
\section{Experiments}

\begin{figure*}[t]
    \centering 
    \includegraphics[width=0.95\textwidth]{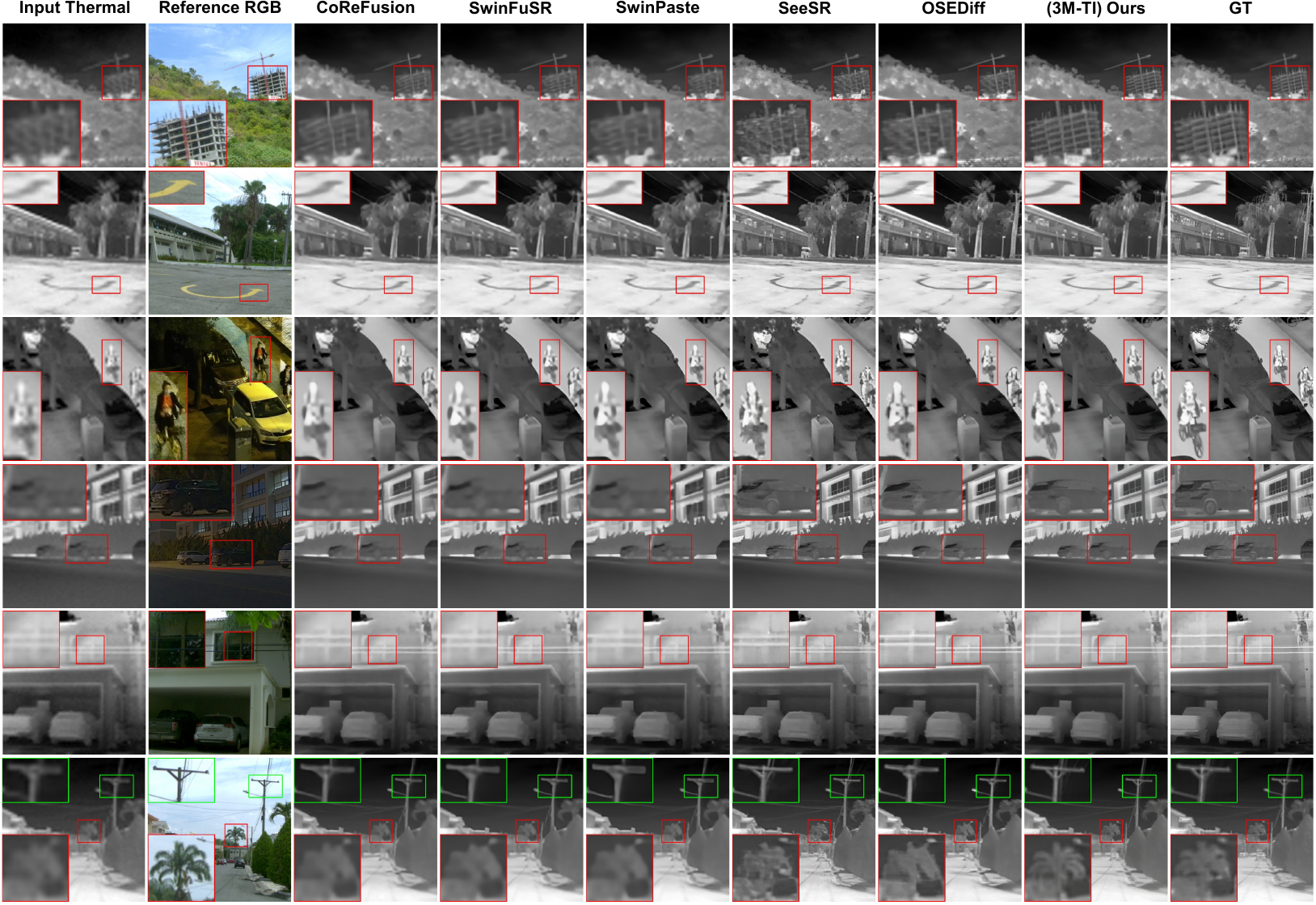}
    \caption{Qualitative comparison on our test set (zoom in for details). 3M-TI achieves the most faithful and visually consistent results, exhibiting sharp structures and accurate thermal patterns that best align with the GT.}
    \label{fig:sota}
\end{figure*}

\begin{figure*}[h]
    \centering 
    \includegraphics[width=0.92\textwidth]{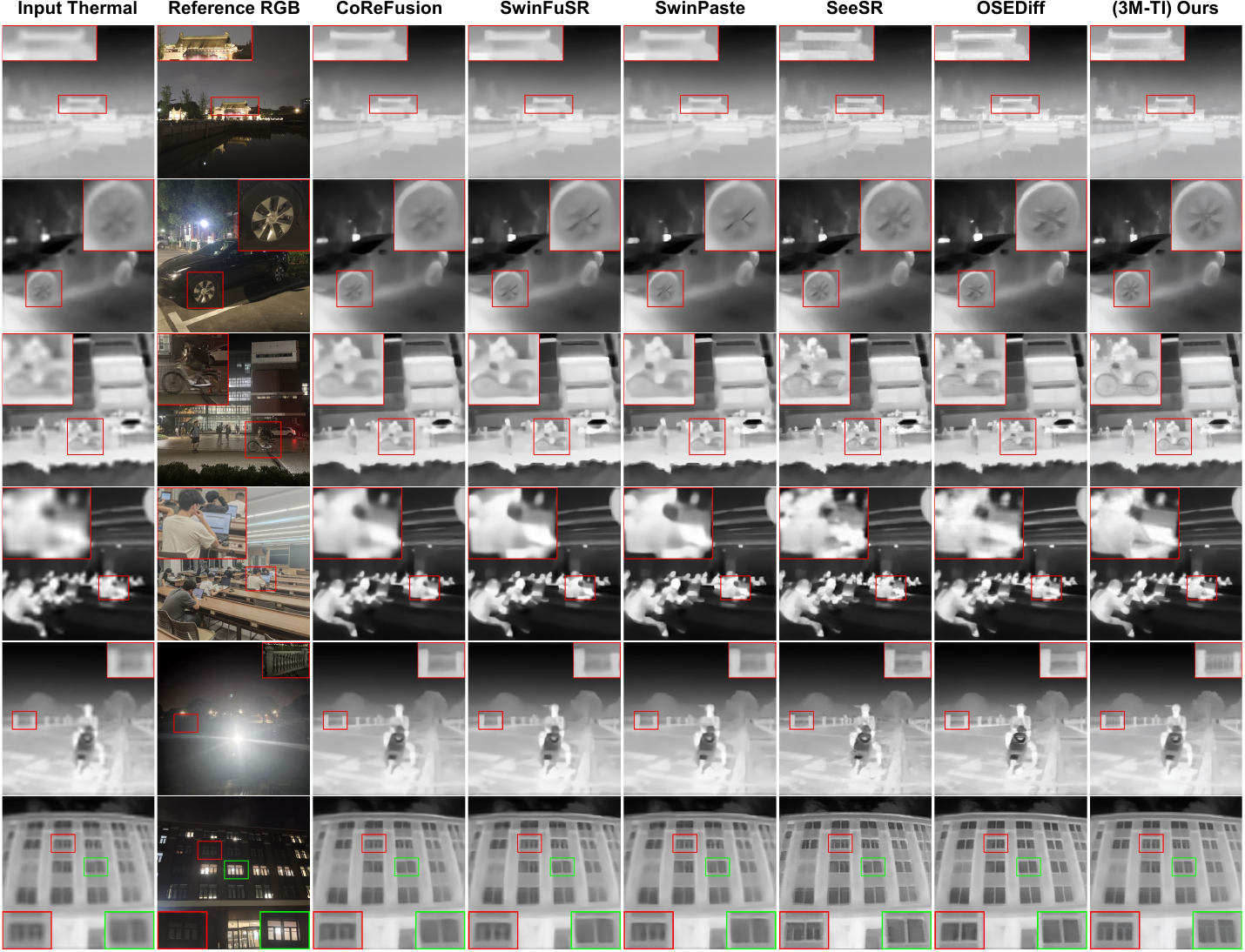}
    \caption{Qualitative comparison on our real-world smartphone dataset (zoom in for details). 3M-TI exhibits remarkable generalization capability, producing sharp and faithful thermal details that closely align with RGB images.}
    \label{fig:real_val}
\end{figure*}

\label{sec:experiment}
\subsection{Experimental Setup}
\textbf{Datasets Preparation.} 
Our training and test sets are compiled from four public RGB-thermal datasets: IRVI~\cite{IRVI}, LLVIP~\cite{LLVIP}, M$^3$FD~\cite{M3FD}, and the PBVS 2025 TISR Challenge Track 2~\cite{PBVS25}. Among them, IRVI, LLVIP, and PBVS 2025 provide official training-validation/test splits, while M$^3$FD does not.
To ensure a balanced and diverse dataset, we sample images from each source as follows: 3,200 pairs from IRVI, 3,200 from LLVIP, 3,822 from M$^3$FD, and 700 from the PBVS 2025 training split, forming a combined training set of 10,922 image pairs. For evaluation, we construct a test set of 1,176 pairs by sampling 300 pairs from the IRVI test set, 300 from LLVIP, 376 from M$^3$FD, and 200 from the PBVS 2025 validation set. 
% A more detailed explanation of datasets sampling is given in the supplementary material.
%
Since the original datasets contain strictly aligned image pairs, we further augment the test set by applying controlled spatial transformations to simulate camera parallax and temporal offset. 
All thermal images are center-cropped to a square aspect ratio and resized to \(64 \times 64\) pixels. To better mimic real sensor noise, we add Gaussian noise. Corresponding RGB images are cropped identically and resized to \(512 \times 512\) pixels to serve as high-resolution references.

\noindent \textbf{Smartphone Dataset.} For real-world evaluation, we collected 300 image pairs across 100 diverse scenes using our smartphone-based multi-camera system, including 200 nighttime and the remaining daytime captures. Similarly, all thermal images are resized to \(64 \times 64\). The corresponding RGB images are center-cropped to a square aspect ratio and resized to \(512 \times 512\) as high-resolution references.

\noindent \textbf{Implementation Details}. 
The model is trained with the loss in Eq.~(\ref{Loss}), combining L2 and learned perceptual image patch similarity (LPIPS)~\cite{lpips} terms with $\lambda=1$: 
\begin{align}
    \mathcal{L} = \mathcal{L}_{2} + \lambda \cdot \mathcal{L}_{\mathrm{LPIPS}}. 
    \label{Loss}
\end{align}
Training is performed using the Adam optimizer~\cite{Adam} with a learning rate of $2\times10^{-5}$ and a batch size of 4 on a single NVIDIA A800 (80~GB) GPU for about 4 hours (8000 iterations). LoRA is applied with ranks of 16 for the UNet and 4 for the VAE decoder.

% The model is trained using the loss in Eq.~(\ref{Loss}), combining L2 and learned perceptual image patch similarity (LPIPS)\cite{lpips} terms, with the weighting factor $\lambda$ set to 1.
% \begin{align}
%     \mathcal{L} = \mathcal{L}_{2} + \lambda \cdot \mathcal{L}_{\mathrm{LPIPS}}. 
%     \label{Loss}
% \end{align}
% We optimize the model using the Adam optimizer~\cite{Adam} with a learning rate of $2\times10^{-5}$ and a batch size of 4 on a single NVIDIA A800 (80~GB) GPU, for approximately 4.5 hours with 8000 iterations. LoRA is applied with rank 16 for the UNet and 4 for the VAE decoder.    

\noindent \textbf{Comparative Methods}. 
We compare our model against 7 representative baselines: CoReFusion~\cite{corefusion}, CoRPLE~\cite{corple}, SwinFuSR~\cite{swinfusr}, SwinPaste~\cite{swinpaste}, SeeSR~\cite{seesr}, OSEDiff~\cite{osediff}, and DifIISR~\cite{difiisr}. 
CoReFusion is an RGB-guided super-resolution model built upon the UNet architecture. 
SwinFuSR employs a Swin Transformer backbone with RGB guidance, while SwinPaste is an enhanced variant. 
SeeSR, OSEDiff, and DifIISR are diffusion-based image super-resolution methods without reference guidance. Among them, DifIISR is specifically designed for infrared images.
CoRPLE is a hybrid framework combining contourlet residual and prompt learning for infrared image enhancement.
For fair comparison, all baseline models except DifIISR are retrained on our dataset using their publicly released codes, while DifIISR is evaluated directly since its training code is unavailable.

\noindent \textbf{Evaluation Metrics}. 
We evaluate model performance using reference metrics, including PSNR and SSIM~\cite{ssim} for reconstruction fidelity and LPIPS~\cite{lpips} for perceptual quality. 
We also report no-reference metrics, MUSIQ~\cite{IQA-pytorch} and MANIQA~\cite{IQA-pytorch}, to quantify overall image quality.
\vspace{-5pt}

\subsection{Image Quality Assesment}
Table~\ref{tab:sota} summarizes the quantitative results. Our 3M-TI achieves the best performance on perceptual metrics, including LPIPS, MANIQA, and MUSIQ. 
For the reference-based LPIPS metric, 3M-TI, OSEDiff, and SeeSR outperform non-diffusion methods (CoReFusion, CoRPLE, SwinFuSR, and SwinPaste), highlighting the power of diffusion priors for perceptual enhancement. In terms of fidelity metrics (PSNR, SSIM), 3M-TI demonstrate better structural preservation than OSEDiff, SeeSR, DifIISR, SwinFuSR, and SwinPaste.
The proposed misalignment augmentation significantly improves 3M-TI, particularly on perceptual metrics, while other methods see only marginal gains. 
This indicates that 3M-TI effectively learns cross-modal correspondences and generalizes well to diverse uncalibrated conditions.

As shown in Fig.~\ref{fig:sota}, qualitative results reveal a key trend: non-diffusion methods tend to achieve high PSNR/SSIM yet produce overly smooth reconstructions that miss fine high-frequency structures. OSEDiff and SeeSR can synthesize high-frequency content but often introduces details that are inconsistent with the ground truth. In contrast, 3M-TI reconstructs fine details that are both visually plausible and closely aligned with the ground truth. 
For example, 3M-TI faithfully transfers geometric details such as partially constructed buildings (Row 1), road markings (Row 2), windows and wires (Row 5), and tree branches and utility poles (Row 6) from the RGB references to the thermal outputs, whereas other methods either miss these structures or reproduce them inaccurately.
% For example, 3M-TI accurately transfers the geometry of partially constructed buildings (Row 1), road markings (Row 2), window and wires (Row 5), tree branches and utility pole (Row 6) from the RGB references to the thermal outputs, while other methods either omit these cues or render them inaccurately. 
3M-TI also yields more natural object contours (e.g., cyclists and cars in the Row 3 and Row 4), resulting in visually realistic thermal images.
% \textcolor{red}{and sharper structures (e.g., window's mullion in the 6th row)}

Figure~\ref{fig:real_val} presents qualitative results on our smartphone dataset, confirming 3M-TI's robustness to real-world, uncalibrated pairs. While CoReFusion, SwinFuSR, and SwinPaste yield blurry outputs; OSEDiff and SeeSR introduce unrealistic artifacts, our method produces geometrically consistent and visually plausible details.
A key example is the bicycle wheels (Row 3), where 3M-TI accurately reconstructs their circular shape despite significant positional misalignment between the RGB and thermal captures due to unsynchronization.
Moreover, 3M-TI can leverage weak cues from degraded references, as shown by its ability to recover the railing structure (Row 5) from an RGB image heavily corrupted by light flare.
Table~\ref{tab:val} reports no-reference evaluation on the captured smartphone dataset, where 3M-TI achieves the highest scores.
More results and technical details are given in the supplementary material.

\vspace{-5pt}
\subsection{Downstream Applications} 
The high-fidelity outputs of 3M-TI provide structurally and semantically rich representations that directly benefit downstream vision tasks. To validate this claim, we evaluate two representative tasks: open-vocabulary object detection and semantic segmentation. In these experiments, we apply the pretrained Grounded-SAM model \cite{GSA} in a zero-shot manner (no fine-tuning) and use identical text prompts for all methods to ensure fair comparisons. 

% Table generated by Excel2LaTeX from sheet 'Sota-Ablation'
\begin{table}[htbp]
  \centering
  \footnotesize
  \setlength{\tabcolsep}{1pt}
    \begin{tabular}{c|ccccc}
    \hline
    \diagbox[width=10em, height=1.8em, innerleftsep=0pt, innerrightsep=0pt]{Methods}{Metrics} & PSNR$\uparrow$ & SSIM$\uparrow$ & LPIPS$\downarrow$ & MANIQA$\uparrow$ & MUSIQ$\uparrow$ \\
    \hline
    CoReFusion \cite{corefusion} & 30.11  & 0.8588  & 0.3214  & 0.2771  & 28.35  \\
    CoReFusion w/ Augment & \cellcolor[rgb]{ .749,  .749,  .749}30.30 & \cellcolor[rgb]{ .749,  .749,  .749}0.8634 & 0.3174  & 0.2748  & 28.95  \\
    \hline
    CoRPLE \cite{corple} & \cellcolor[rgb]{ .502,  .502,  .502}\textbf{30.47} & \cellcolor[rgb]{ .502,  .502,  .502}\textbf{0.8642} & 0.3206 & 0.2833 & 30.46 \\
    \hline
    SwinFuSR \cite{swinfusr} & 29.85  & 0.8549  & 0.3085  & 0.2740  & 29.86  \\
    SwinFuSR w/ Augment & 29.98  & 0.8581  & 0.3134  & 0.2753  & 30.33  \\
    \hline
    SwinPaste \cite{swinpaste} & 29.83  & 0.8545  & 0.3075  & 0.2719  & 29.63  \\
    SwinPaste w/ Augment & 29.91  & 0.8575  & 0.3084  & 0.2745  & 30.66  \\
    \hline
    SeeSR \cite{seesr} & 29.41  & 0.8495  & \cellcolor[rgb]{ .749,  .749,  .749}0.1828  & \cellcolor[rgb]{ .749,  .749,  .749}0.4278  & 35.22  \\
    \hline
    OSEDiff \cite{osediff} & 28.05  & 0.8422  & 0.2113  & 0.4014  & 36.30 \\
    \hline
    DifIISR \cite{difiisr} & 27.48 & 0.7905 & 0.3484 & 0.4214 & \cellcolor[rgb]{ .502,  .502,  .502}\textbf{36.74} \\
    \hline
    Ours w/o Augment & 30.04  & 0.8597  & 0.1917  & 0.4265  & 34.94  \\
    Ours w/ Augment & 30.09  & 0.8610  & \cellcolor[rgb]{ .502,  .502,  .502}\textbf{0.1787 } & \cellcolor[rgb]{ .502,  .502,  .502}\textbf{0.4443} & \cellcolor[rgb]{ .749,  .749,  .749}36.66 \\
    \hline
    \end{tabular}%
    \caption{Performance comparison of different methods on public datasets. Gray cells indicate the best result, and light gray cells indicate the second-best result for each metric.}
  \label{tab:sota}%
  \vspace{-15pt}
\end{table}%

% Table generated by Excel2LaTeX from sheet 'Val'
\begin{table}[htbp]
  \centering
  \scriptsize
  \setlength{\tabcolsep}{1.5pt}
    \begin{tabular}{c|cccccc}
    \hline
    \diagbox{Metrics}{Methods} & CoReFusion & SwinFuSR & SwinPaste & SeeSR & OSEDiff & 3M-TI \\
    \hline
    MUSIQ$\uparrow$ & 25.74  & 25.99  & 26.17 & 30.15 & 29.85  & \cellcolor[rgb]{ .502,  .502,  .502}\textbf{30.62} \\
    MANIQA$\uparrow$ & 0.2701  & 0.2754  & 0.2749 & 0.3454 & 0.3285  & \cellcolor[rgb]{ .502,  .502,  .502}\textbf{0.3589} \\
    \hline
    \end{tabular}%
    \caption{Performance comparison on real-world smartphone dataset. Gray cells indicate the best.}
  \label{tab:val}%
  \vspace{-15pt}
\end{table}%

\begin{figure}[t]
    \centering
    \includegraphics[width=0.9\columnwidth]{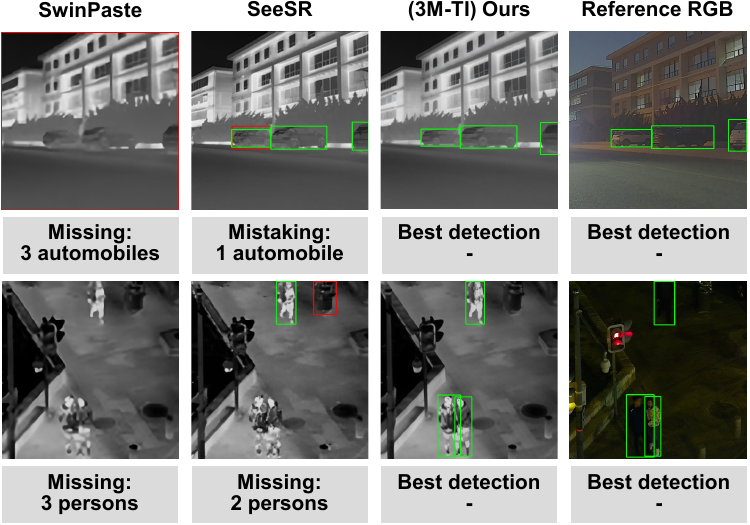}
    \caption{Visualization of detection results, where green bounding boxes indicate the correct detection, red bounding boxes indicate the wrong detection.}
    \vspace{-15pt}
    \label{fig:dectect}
\end{figure}

\noindent \textbf{Object detection.} For the examples in Fig.~\ref{fig:dectect}, the text prompts are ``automobile'' and ``person''. 3M-TI demonstrates superior detection robustness and object discovery capability compared to SwinPaste and SeeSR. Table~\ref{tab:detect} summarizes detection performance (Precision, Recall, F1-score, and IoU) on the test set (LLVIP and M$^{3}$FD with GT annotations). 3M-TI achieves the best overall detection performance, marginally surpassing the reference RGB results and closely approaching the scores obtained on GT thermal images. 
% Furthermore, the reconstructed thermal and RGB detection exhibit complementary behavior. In the second row, the RGB image is able to separate adjacent persons but suffers from false positives, while the thermal reconstruction produces no false alarms but tends to merge closely spaced persons.

\begin{figure}
    \centering
    \includegraphics[width=0.95\columnwidth]{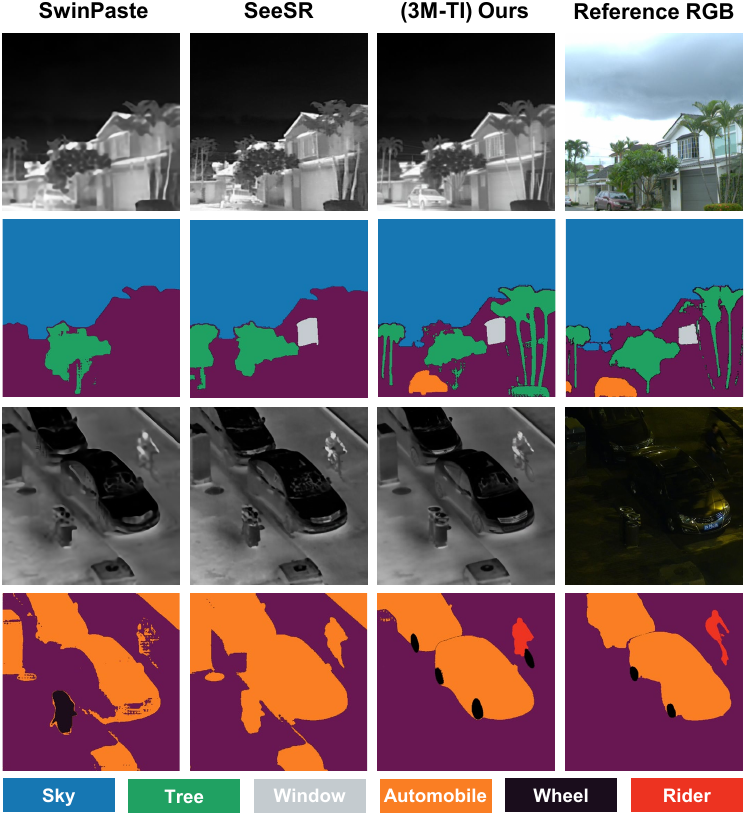}
    \caption{Visualization of segmentation results, where different colors represent different object categories.}
    \label{fig:segment}
    \vspace{-10pt}
\end{figure}

\noindent \textbf{Semantic segmentation.} For the examples in Fig.~\ref{fig:segment}, the prompts are ``sky, tree, automobile, window'' and ``wheel, automobile, rider''. 3M-TI produces more accurate and coherent segmentation maps than other methods, while complementing the RGB reference. In the first example, compared with the RGB results, 3M-TI segments trees more precisely though it misses a small automobile in the bottom-left. In the second low-light example, 3M-TI even outperforms the RGB reference (e.g., wheels and the upper automobile). 

Overall, these results confirm that 3M-TI generates semantically meaningful details that enhance performance in critical vision tasks, demonstrating its practical utility beyond perceptual quality.

% Table generated by Excel2LaTeX from sheet 'detection'
\begin{table}[htbp]
  \centering
  \footnotesize
  \setlength{\tabcolsep}{4pt}
    \begin{tabular}{c|cccc}
    \hline
    \diagbox{Methods}{Metrics} & Precision$\uparrow$ & Recall$\uparrow$ & F1-score$\uparrow$ & IoU$\uparrow$ \\
    \hline
    Thermal (SwinPaste) & 0.1800  & 0.2109  & 0.1765  & 0.1941  \\
    Thermal (SeeSR) & 0.3832  & 0.4637  & 0.3849  & 0.3022  \\
    Thermal (3M-TI)  & \cellcolor[rgb]{ .502,  .502,  .502}\textbf{0.4565} & \cellcolor[rgb]{ .749,  .749,  .749}0.5455  & \cellcolor[rgb]{ .502,  .502,  .502}\textbf{0.4724} & \cellcolor[rgb]{ .502,  .502,  .502}\textbf{0.3427} \\
    Reference RGB   & \cellcolor[rgb]{ .749,  .749,  .749}0.4322  & \cellcolor[rgb]{ .502,  .502,  .502}\textbf{0.5708} & \cellcolor[rgb]{ .749,  .749,  .749}0.4643  & \cellcolor[rgb]{ .749,  .749,  .749}0.3359  \\
    \hline
    Thermal (GT)  & 0.4582 & 0.5793  & 0.4887 & 0.3494 \\
    \hline
    \end{tabular}%
  \caption{Detection performance comparison across different methods, reference RGB, and GT, evaluated by Precision, Recall, F1-score, and IoU.}
  \label{tab:detect}%
  \vspace{-15pt}
\end{table}%

\begin{table}[htbp]
  \centering
  \footnotesize
    \begin{tabular}{c|cccc}
    \hline
    \diagbox{Methods}{Metrics} & PSNR$\uparrow$ & SSIM$\uparrow$ & LPIPS$\downarrow$ & MUSIQ$\uparrow$ \\
    \hline
    w/o Reference & 29.97  & 0.8592  & 0.2106  & 32.85  \\
    w/o RAM prompt & 30.09  & \cellcolor[rgb]{ .502,  .502,  .502}\textbf{0.8616}  & 0.1805  & 36.20 \\
    w/o Augmentation & 30.04  & 0.8597  & 0.1917  & 34.94  \\
    w/o Skip & 29.86  & 0.8572  & 0.1795  & 36.58 \\
    \hline
    w/ Self-Attn & 30.03  & 0.8591  & 0.1954  & 34.47 \\
    w/ Feature Concat & 29.60  & 0.8481  & 0.2164  & 32.12 \\
    w/ Cross-Attn & 29.86  & 0.8555  & 0.1953  & 34.32 \\
    \hline
    w/ All (w/ CSM)  & \cellcolor[rgb]{ .502,  .502,  .502}\textbf{30.09 } & 0.8610 & \cellcolor[rgb]{ .502,  .502,  .502}\textbf{0.1787 } & \cellcolor[rgb]{ .502,  .502,  .502}\textbf{36.66}  \\
    \hline
    \end{tabular}
  \caption{Ablation study of 3M-TI components. Gray cells indicate the best result for each metric.}
  \label{tab:ablation}
  \vspace{-10pt}
\end{table}

% \begin{figure}
%     \centering
%     \includegraphics[width=0.95\columnwidth]{sec/ablation.pdf}
%     \caption{Qualitative results of the ablation study.}
%     \label{fig:ablation}
%     \vspace{-10pt}
% \end{figure}

\vspace{-5pt}
\subsection{Ablation Study}
We conduct an ablation study to investigate the contributions of each component. Table~\ref{tab:ablation} and Fig. S5 in the supplementary material report quantitative drops and representative visual examples when individual modules are removed. Removing the RGB reference produces markedly blurrier reconstructions (e.g., bicycle spokes and shrub textures), indicating the importance of cross-modal cues. Removing the misalignment augmentation reduces robustness to geometric and temporal offsets, leading to visibly degraded high-frequency details. Removing the skip connection degrades structural fidelity: circular wheels become distorted and geometric consistency is lost. Removing the semantic guidance of RAM hardly impairs reference metrics yet brings a drop in perceptual quality. With all components enabled (w/ All), 3M-TI attains the best balance of fidelity and perceptual quality, producing sharp, semantically consistent thermal reconstructions.

We further compared CSM (w/ CSM) against: 1) Original Self-Attn (intra-modal attention only); 2) Feature Concat + FC (static projection w/o attention); 3) Standard Cross-Attn (inter-modal attention only).
The lower half of Table \ref{tab:ablation} show that CSM outperforms all variants. Original Self-Attn is incapable of fusing RGB cues. Feature Concat relies on static weights, lacking content-adaptive capabilities. Standard Cross-Attn neglects the internal spatial context within the thermal modality itself. In contrast, CSM concatenates tokens to perform joint Self-Attn. This enables the model to capture both inter-modal (RGB-thermal) guidance and intra-modal (thermal-thermal) structural dependencies, which is vital for integrating complementary RGB details while preserving thermal structures.

% Table generated by Excel2LaTeX from sheet 'Sota-Ablation'

%% file: sec/5_conclusion.tex
\vspace{-10pt}
\section{Conclusion}
\label{sec:conclusion}
\vspace{-5pt}
We proposed 3M-TI, a cross-modal diffusion framework that integrates calibration-free RGB guidance to enhance thermal image reconstruction. 
With the cross-modal self-attention module and misalignment augmentation, 3M-TI effectively aligns and fuses information across modalities, producing thermal images with sharper, more realistic, and semantically consistent details. 
Comprehensive experiments and real-system evaluations demonstrate that 3M-TI not only achieves superior perceptual quality but also significantly benefits downstream machine vision tasks.

% We proposed a calibration-free multi-camera cross-modality diffusion framework for thermal image super-resolution. we introduced an alignment-oriented augmentation strategy to simulate multi-camera parallax and unsynchronized captures. A one-step diffusion model with CSA layers were trained on the augmented datasets to adaptively capture correspondence and learn feature fusion under uncalibrated modalities. Experimental results on both synthetic datasets and real-world images demonstrated that our method outperformed existing state-of-the-art approaches, achieving superior visual and perceptual performance. When employing the images generated by our method for detection and segmentation, Grounded-SAM exhibited higher accuracy.

\section*{Acknowledgments}
This work is supported in part by the National Natural Science Foundation of China (NSFC) under Grant No. 62271283, and in part by the Shanghai Jiao Tong University Medical-Engineering Cross Research Fund under Grant No. YG2026QNB50.